# Moving Healthcare AI-Support Systems for Visually Detectable Diseases onto Constrained Devices

Tess Watt, Christos Chrysoulas, Peter J Barclay
*School of Computing, Engineering & the Built Environment*
*Edinburgh Napier University*
Edinburgh, Scotland
{t.watt, c.chrysoulas, p.barclay}@napier.ac.uk

*Abstract*—Image classification usually requires connectivity and access to the cloud which is often limited in many parts of the world, including hard to reach rural areas. TinyML aims to solve this problem by hosting AI assistants on constrained devices, eliminating connectivity issues by processing data within the device itself, without internet or cloud access. This pilot study explores the use of tinyML to provide healthcare support with low spec devices in low connectivity environments, focusing on diagnosis of skin diseases and the ethical use of AI assistants in a healthcare setting. To investigate this, 10,000 images of skin lesions were used to train a model for classifying visually detectable diseases (VDDs). The model weights were then offloaded to a Raspberry Pi with a webcam attached, to be used for the classification of skin lesions without internet access. It was found that the developed prototype achieved a test accuracy of 78% and a test loss of 1.08.

*Keywords—tinyML, computation offloading, artificial intelligence, machine learning, computer vision, image classification*

## I. INTRODUCTION

Artificial intelligence (AI) assistants are extremely useful and important in the healthcare domain. However, AI assistants currently require an internet connection to function, which limits their use-cases. This is an issue, as a report by the United Nations (UN) found that an estimated 37% of the world's population have never used the internet [1]. Being unable to connect to the internet, or having intermittent access, means being unable to access data from the cloud. Constrained devices (also known as 'low resource' devices) are designed to work in such an environment. This presents an exciting opportunity to make healthcare AI-support systems accessible to those in low-connectivity areas. The ability to deploy healthcare AI-support systems on constrained devices could allow for them to be distributed in remote areas and extreme environments such as rural areas, caves, mines, and even space. One way to achieve this is using tinyML (tiny machine learning), which makes use of small devices such as microcontrollers for the deployment of machine learning (ML) algorithms.

In recent years there have been many advancements in ML, computer vision, and image classification for use in the medical domain. Deep learning, particularly the use of Convolutional Neural Networks (CNNs), has become the state-of-the-art for automated visual understanding [2]. As stated in a 2023 paper by Huang et al. [3], deep learning shows promise for medical image analysis, which could in turn improve patient outcomes. This is imperative as due to the rapid increase in the number of medical images produced, medical professionals are struggling to keep up with the workload [3]. Developing AI systems that can automate arduous and time-consuming processes such as diagnosing VDDs could help to solve this problem.

To address this problem, we build on the recent advances in image classification and computer vision with the following contributions:

1) We present the architecture design for a tinyML prototype based on device capacity and state-of-the-art image classification approaches that can be used to identify certain diseases by pointing the device camera at the patient.

2) We introduce a novel application of image classification for low-resource situations.

3) We provide experimental evaluation and comparative assessment of our approach against existing image classification models.

The rest of the paper is structured as follows. Section II reviews related work. Section III discusses ethical considerations and Section IV describes our proposed system architecture and implementation. Section V provides a discussion of the results and comparative assessment, and finally Section VI concludes our work.

## II. RELATED WORK

This section presents a discussion of the related cloud computing and tinyML, focusing on image classification.

### A. Cloud Computing

According to Guo et al. [4], cloud computing is one of the three major developments the internet has experienced since 2010. The other two are the growing expansion of mobile devices, and the Internet of Things (IoT). Cloud computing has been defined as a universal computing model for enabling on-demand access to a shared pool of resources [4]. Typically, cloud computing is carried out via remote, centralized data centres. Some of the most well-known cloud services are Amazon Web Services (AWS), Microsoft Azure, and Google Cloud [5]. However, this traditional approach to cloud computing is becoming increasingly unable to handle the massive distribution of big data and the processing power required to analyse it [6].

The challenges that centralized cloud computing faces were explored by Ferrer et al. [7]. Within the current cloud computing infrastructure, resources are treated as effectively infinite. Despite the perceived unlimited computing resources, centralized cloud environments are experiencing serious problems with large data movements and latency, especially when it comes to IoT [8]. Furthermore, Tambe et al. [9] claim that the IoT market is being driven by the rapid increase in the number of AI assistants. AI assistants typically require sophisticated language models with a vast number of parameters and a high demand for memory [10], which in turn is making power consumption a prominent issue [7]. Considering this, and the latency issues cloud has in IoT scenarios, the cloud may not be the best place to host AI assistants, and especially not for use in remote environments.

## B. TinyML

Being unable to connect to the internet, or having intermittent access, means being unable to access data from the cloud. TinyML and edge computing could solve this issue, as hosting AI assistants on constrained devices can eliminate connectivity issues by processing data within the device itself, without internet or cloud access [11].

However, as AI applications such as healthcare AI-support systems are memory-intensive and require data to be processed quickly, it is important to examine the capability of the constrained device(s) chosen to deploy them. Table I shows a variety of easily available devices that could be used to host a healthcare AI assistant.

TABLE I. STORAGE CAPACITIES AND PROCESSOR SPEEDS OF CONSTRAINED DEVICES

| Device | Memory | Speed |
|---|---|---|
| Raspberry Pi 2 Model B | 1GB | 900MHz |
| Raspberry Pi 3 Model B | 1GB | 1.2GHz |
| Raspberry P 4 Model B | Up to 8GB | 2.5GHz |
| Xilinx PYNQ-Z1 | 650MB | 525MHz |
| UDOO BOLT V3 | Up to 32GB | 2.3GHz |
| Orange Pi 5 | 8GB (up to 32GB) | 2.4GHz |
| Nvidia Jetson Nano | 4GB | 1.43GHz |

From this table the most well-known and widely used devices are the Raspberry Pis. They are also the most affordable options, costing approximately £20-£40 at the time of writing. However, devices such as the UDOO BOLT and Orange Pi are advantageous as the larger memory and fast processor speeds that come with those boards make them suitable for AI and image processing applications. Additionally, the Nvidia Jetson Nano, Orange Pi 5, and UDOO BOLT come with GPU acceleration, which is crucial for running real-time inference. However, these benefits come at a price, with the three aforementioned devices costing at least three times more than Raspberry Pis.

Miori et al. [12] and Javed et al. [13] conducted research into using a cluster of Raspberry Pis to host serverless platforms. Miori's work [12] found that it was possible, but that there were shortcomings in performance compared to standard PC hardware. However, as this research was published in 2017, Raspberry Pi 2 models were used; this may have contributed to the performance issues found. Despite this, a positive outcome of their research is that they concluded that with the next generation of devices to come (in this case the Raspberry Pi 3 model and beyond), successfully hosting serverless platforms will be more viable, especially for use-cases in harsh environments.

Cost is another benefit of using low-resource devices instead of expensive cloud servers. This is supported by G. Liu et al. [14] who state that one of the reasons edge computing has attracted so much interest over recent years is because of its low communication costs. These communication costs are reduced owing to the ability of many edge nodes to pre-process large amounts of data before it is sent to the cloud. Even higher-end devices such as the UDOO BOLT, which costs around £450, are still cheaper in the long-term than the cost of using a cloud server, as they are a one-time purchase rather than a monthly fee. Additionally, using tinyML rather than the cloud to deploy AI assistants reduces the $CO_2$ footprint and energy consumption of running pre-trained machine learning algorithms [11].

In recent years, there have been developments in tinyML by IBM [15]. IBM's Watson Assistant is a flexible conversational AI platform that can be employed for a variety of use-cases, one of them being a healthcare chatbot. This has been used to create an edge computing robot interface for mental health care [16]. Users interacted with the system by responding to specialist questions, asked via audio, and answered via text. The system was developed on a Xilinx PYNQ-Z1 board (see Table I), which was used as an embedded device that can be placed anywhere within the home. It was programmed in Python and operates using a combination of libraries from Google Cloud and IBM Watson. The processing speed of the PYNQ board was evaluated against a PC using an I5 processor, and it was found that in the worst case, the PYNQ board spent 21 seconds per question compared to 37 seconds by the PC. From the user interacting with this system, they were able to receive mental health diagnoses after the results were verified by clinical follow-up. More discussion of AI-produced diagnoses and medical verification appears in Sections C and D.

## C. Image Classification

Image classification for the detection of skin lesions with higher spec computers is widely discussed in the literature. Qin et al. [17] developed an image classification model to classify skin lesions, including melanoma and dermatofibroma, using a pretrained deep neural network and transfer learning. Their data was synthetically generated using a generative adversarial network (GAN). The classifier trained for 50 epochs and achieved an accuracy of 95.2%. Their setup consisted of a computer with 192GB RAM and an Nvidia GPU.

Additionally, Benyahia et al. [18] present extremely in-depth research into the efficiency of 17 CNNs and 24 ML classifiers for the classification of skin lesions. They tested these on two different datasets: ISIC 2019[1] and PH2[2]. It was found that DenseNet201 [19] combined with Fine KNN or Cubic SVM achieved the highest accuracy (92.3%) on the ISIC dataset, and an accuracy of 99% on the PH2 dataset. To run their experiments, they used a desktop computer featuring a Core i9 processor, 32GB DDRAM, and an Nvidia GPU.

However, there are fewer sources discussing these kinds of systems being deployed on constrained/low-spec devices without GPUs. Ramlakhan & Shang [20] developed a mobile melanoma recognition system using a k-nearest neighbors (KNN) classifier [21]. The system achieved an average accuracy of 66.7%, and it was trained on a Nexus One smartphone, which has a 1GHz processor and 512MB RAM. As can be seen, Ramlakhan & Shang's system had a much lower accuracy than the systems that ran on higher spec

---
[1] https://challenge.isic-archive.com/data/#2019

[2] ADDI - Automatic_computer-based_Diagnosis_system_for_Dermoscopy_Images (up.pt)

devices. This was most likely due to the lower spec hardware used.

## III. ETHICAL CONSIDERATIONS

The ethical use of AI assistants in a healthcare setting is discussed in the literature with four main points of focus: collection and storage of sensitive data, impacts of diagnoses on users, medical verification and legal approval of medical devices, and bias and diversity within AI.

Ethically collecting and storing healthcare data is discussed by Xu et al. [22]. The risk of stigmatization or discrimination to users if their data is wrongfully disclosed makes health data especially sensitive. The work of [22] also raised the point that diagnostic predictions given by chatbots can lack transparency, and how this can impact users. This is an example of the "Black Box" problem. This lack of transparency concerning diagnostic inference could potentially be lessened using explainable AI [23]. Additionally, with the recent rise in chatbots such as ChatGPT, their reliability has been scrutinized. Chatbots frequently produce highly convincing statements that are verifiably false [24]. If a user were to undergo an incorrect course of treatment before confirming a predicted diagnosis with a medical professional, the outcome could be fatal. However, the long-term impacts that chatbots could have in healthcare, a vastly complex sociotechnical system, are currently unknown [25].

As an attempt to regulate and deter healthcare chatbots from producing inaccurate diagnoses, EU and US law require that software tools used for diagnosis are approved as medical devices before being put on the market [24]. Another important ethical consideration is potential bias and lack of diversity within AI. Determining the source(s) of potential bias internally within a system (training data) and externally (developers of that system) could help enable systems to better reflect those they are designed to benefit [22]. Ensuring that both the developers of a system, and the data it is trained on, are diverse in terms of gender, race, socioeconomic background, and other factors, could positively influence the health of its users [25]. This is especially important as some diseases disproportionally affect one sex or racial group, which could be under-represented or missed in training data. Specifically, VDDs such as skin lesions could look quite different against different skin coloration.

## IV. SYSTEM ARCHITECTURE

Our main objectives for developing a healthcare AI-support system were to (1) design a tinyML prototype based on device capacity and state-of-the-art image classification approaches, and (2) introduce a novel application of image classification for low-resource situations. A key consideration for the architecture design of our proposed healthcare AI-support system was the storage capacity and processor speed of the constrained device. It was important to examine the capabilities of various devices that could be used to deploy the image classification model. Table 1 shows the devices that were considered. As image classification models are memory-intensive, require data to be processed quickly, and need to be affordable for distribution in low-connectivity areas, a Raspberry Pi 3 Model B was chosen as the device for prototyping. Fig. 1 gives a high-level overview of the proposed system.

The HAM10000 dataset[3] was used to train the system. This dataset consists of 10,000 images of common pigmented skin lesions. The seven classes of skin lesions included in the dataset are benign keratosis, melanocytic nevus, dermatofibroma, melanoma, vascular lesion, basal cell carcinoma, and actinic keratosis. A sample of the data can be seen in Fig. 2.

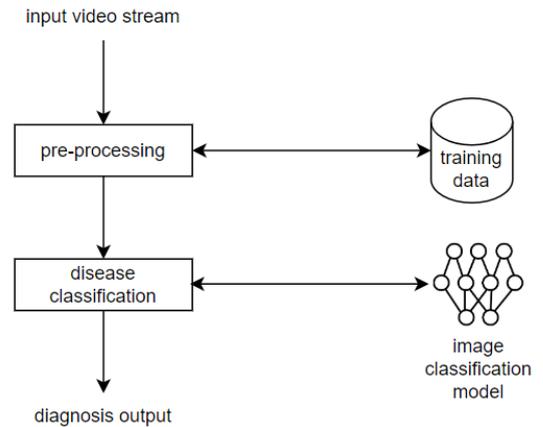

Fig. 1. Healthcare AI-support system component architecture.

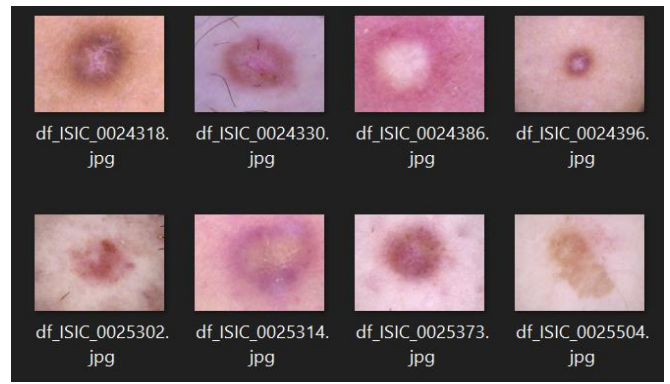

Fig. 2. Sample of dermatofibroma images from the HAM10000 dataset.

The model was trained for ten epochs on a High-Performance Computer (HPC), specifically a HP Z4, and was then offloaded to a Raspberry Pi for use. The webcam used had a resolution of 1080p. Our code and setup instructions are available on GitHub[4]. The model was trained using a CNN, specifically the MobileNet-V2 architecture [26]. Google's MediaPipe API[5] was used to achieve the bridge between the image classification model and the real-time video input. This architecture performed the best out of the four available for use with the MediaPipe API as shown in Table II. [27]

---

[3] https://dataverse.harvard.edu/dataset.xhtml?persistentId=doi:10.7910/DVN/DBW86T
[4] https://github.com/Tess314/Healthcare_AI_Support_System
[5] https://developers.googleblog.com/2023/08/mediapipe-for-raspberry-pi-and-ios.html?m=1

TABLE II. RESULTS OF AVAILABLE MODEL ARCHITECTURES

| Model Architecture | Test Accuracy (%) | F1 Score (%) | Precision (%) | Recall (%) |
|---|---|---|---|---|
| MobileNet-V2 | 78 | 41 | 87 | 71 |
| EfficientNet-Lite0 | 73 | 32 | 83 | 65 |
| EfficientNet-Lite2 | 73 | 30 | 86 | 64 |
| EfficientNet-Lite4 | 71 | 31 | 85 | 61 |

To demonstrate how the system could be used on real patients, we pointed the system's webcam at skin lesion images from the UK National Health Service (NHS) website, and the probability scores for which class of skin lesion the system predicted for each image were displayed. The user interface (UI) can be seen in Fig. 3. and the probability scores for classifying each skin lesion can be seen Table III. This demonstrates the usage scenario and lays the foundation for testing in a realistic environment in the future.

TABLE III. PROBABILITY SCORES FOR CLASSIFYING SKIN LESIONS

| True label | Prob. Scores (%) |
|---|---|
| benign keratosis | benign keratosis – 35<br>basal cell carcinoma – 32<br>melanocytic nevus – 10<br>dermatofibroma – 9<br>actinic keratosis - 8 |
| melanocytic nevus | melanocytic nevus – 51<br>vascular lesion – 12<br>melanoma – 12<br>benign keratosis – 9<br>actinic keratosis - 9 |
| dermatofibroma | melanocytic nevus – 67<br>vascular lesion – 12<br>dermatofibroma – 10<br>melanoma – 5<br>actinic keratosis - 3 |
| melanoma | melanoma – 95<br>melanocytic nevus – 2<br>benign keratosis – 1<br>dermatofibroma -1<br>vascular lesion -1 |
| vascular lesion | benign keratosis – 47<br>melanocytic nevus – 28<br>melanoma – 13<br>vascular lesion – 4<br>dermatofibroma 4 |
| basal cell carcinoma | melanoma – 53<br>vascular lesion – 14<br>basal cell carcinoma – 10<br>benign keratosis – 9<br>actinic keratosis - 8 |
| actinic keratosis | actinic keratosis – 33<br>benign keratosis – 27<br>melanoma – 14<br>basal cell carcinoma – 8<br>melanocytic nevus - 7 |

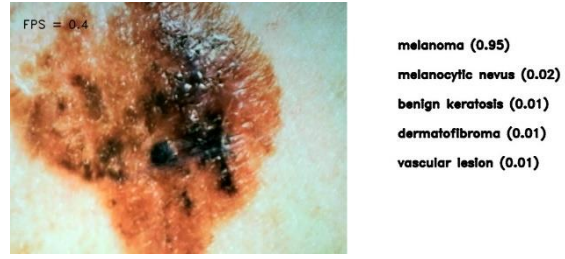

Fig. 3. Probability scores for classifying melanoma.

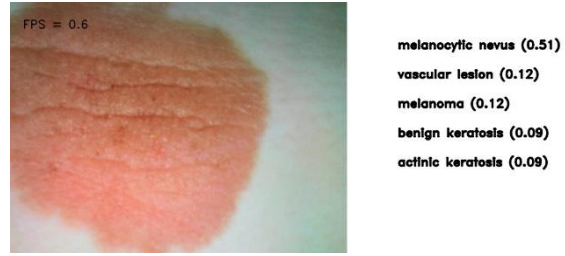

Fig. 4. Probability scores for classifying melanocytic nevus.

V. EVALUATION AND RESULTS

The model produced a test accuracy of 78% and a test loss of 1.08 The test data consisted of 1000 images. In Fig. 5., we observe a consistent improvement in both training and validation accuracy with each epoch, along with a consistent decrease in loss values. This suggests that the model did not suffer from overfitting. However, it is worth noting that in our first experiments neither the accuracy nor the loss is particularly impressive. This will be addressed in our future work comparing and tuning different algorithms.

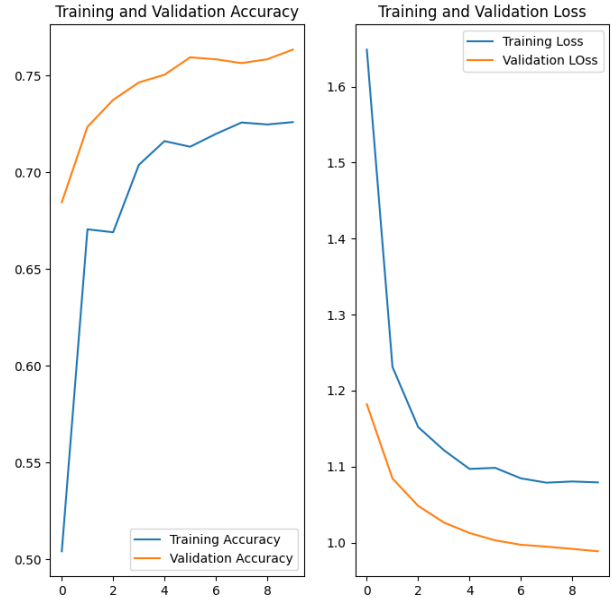

Fig. 5. Model accuracy and loss vs epoch

To further evaluate the model, its architecture, parameters, and evaluation metrics, we compare our results with those coming from the literature (see Table IV). We only consider the accuracy here as F1 score, precision, and recall were not reported in all papers.

TABLE IV. COMPARISON OF IMIAGE CLASSIFICATION MODEL METRICS

| Model | Accuracy (%) |
|---|---|
| Benyahia et al. | 99.0 |
| Qin et al. | 95.2 |
| Ramlakhan & Shang | 66.7 |
| Proposed Model | 78.0 |

As shown, our proposed model produced a higher accuracy than the model developed by Ramlakhan & Shang [20] but produced a lower accuracy than the other two models. One potential reason our model obtained a higher accuracy than that of Ramlakhan & Shang is because they used a smaller dataset (83 lesion images). Qin et al. used the ISIC 2018 dataset, which consists of 10,015 images. As this is similar to the size of the HAM10000 dataset we used, it is evident that they have a higher accuracy due to factors such as the algorithm they used and how it was fine-tuned. Similarly, Benyahia et al. used the PH2 dataset which only contains 200 images, meaning their higher accuracy may also come down to more specific factors.

## VI. CONCLUSION

An image classification system for identifying VDDs was deployed on a constrained device. This was done as a pilot study for systems to be used in remote areas, and to avoid the connectivity issues associated with cloud computing in IoT scenarios. We reviewed the most appropriate hardware to use and selected the Raspberry Pi 3 for our study. The identified software components necessary for developing the image classification system were presented, and the system was evaluated against the results of existing models.

Future work will include comparing a wider range of algorithms and investigating tuning the most promising candidates, as well as using model compression techniques such as quantization and pruning. Additionally, more robust evaluation methods and additional evaluation metrics will be considered to validate the results. More powerful boards such as the UDOO BOLT and Orange Pi 5 could also play a significant role in expanding and improving this work. Edge computing could also be integrated for the use of federated learning. With these improvements in place, this system could pave the way for low-spec devices to be used for real-time detection of VDDs in remote and low-connectivity areas.